\documentclass{article}
\usepackage[utf8]{inputenc}
\usepackage{amsmath,amssymb}
\usepackage{booktabs}
\usepackage{graphicx}
\usepackage{comment}
\usepackage[margin=1in]{geometry}
\usepackage{hyperref}

\author{
  Ayyüce Begüm Bektaş$^{1*}$ \quad
  Mithat Gönen$^{1}$ \\[0.5em]
}

\title{Machine Learning for Medicine Must Be Interpretable, Shareable, Reproducible and Accountable by Design}

\date{}

\begin{document}
\maketitle

\begingroup
\renewcommand\thefootnote{}\footnotetext{\textsuperscript{1} Sloan Kettering Institute for Cancer Research, New York, NY, USA.}
\footnotetext{\textsuperscript{*} Corresponding author: bektasa@mskcc.org}
\endgroup

\vspace*{-1em}
\begin{abstract}
This paper claims that machine learning models deployed in high stakes domains such as medicine must be interpretable, shareable, reproducible and accountable. We argue that these principles should form the foundational design criteria for machine learning algorithms dealing with critical medical data, including survival analysis and risk prediction tasks. Black box models, while often highly accurate, struggle to gain trust and regulatory approval in health care due to a lack of transparency. We discuss how intrinsically interpretable modeling approaches (such as kernel methods with sparsity, prototype-based learning, and deep kernel models) can serve as powerful alternatives to opaque deep networks, providing insight into biomedical predictions. We then examine accountability in model development, calling for rigorous evaluation, fairness, and uncertainty quantification to ensure models reliably support clinical decisions. Finally, we explore how generative AI and collaborative learning paradigms (such as federated learning and diffusion-based data synthesis) enable reproducible research and cross-institutional integration of heterogeneous biomedical data without compromising privacy, hence shareability. By rethinking machine learning foundations along these axes, we can develop medical AI that is not only accurate but also transparent, trustworthy, and translatable to real-world clinical settings.
\end{abstract}

\section{Introduction}

\textbf{This paper argues that machine learning models for medicine must be interpretable, shareable, reproducible, and accountable by design.} These properties should not be treated as optional refinements or secondary considerations, but as foundational criteria for any system intended to support real-world clinical decision-making.

Modern machine learning has demonstrated remarkable capabilities in analyzing complex biomedical data, from genomic profiles to electronic health records. Large-scale resources such as The Cancer Genome Atlas (TCGA)~\cite{cancergenomeatlas2013pan}, the MIMIC-IV intensive care database~\cite{johnson2023mimic}, and the UK Biobank~\cite{ref3} exemplify the opportunities: they provide rich datasets for developing models to predict patient outcomes, survival, and treatment responses. However, models deployed for high-stakes decisions—such as predicting cancer patient survival or guiding therapy—must meet a higher standard than those used in low-risk or exploratory domains.

In particular, clinicians and regulators increasingly demand that such models be interpretable (so human experts can understand and trust the reasoning), shareable (so knowledge can be disseminated across institutions despite privacy constraints), reproducible (so results can be independently validated), and accountable (so performance and decision behavior can be audited and scrutinized). Despite significant progress in deep learning, many state-of-the-art models remain black boxes with millions of parameters and no intrinsic transparency. In clinical settings, this opacity often prevents trust and adoption, since physicians are unlikely to rely on models that cannot justify their outputs. It also raises safety concerns when model behavior generalizes poorly across populations or institutions.

Recent work has argued that, in high-stakes applications, inherently interpretable models should be favored over post hoc explanations of black-box systems~\cite{rudin2019stop}. We endorse this view and contend that interpretability must be treated not as an afterthought but as a core design principle in biomedical ML. Interpretability supports accountability (by enabling auditability and debugging), facilitates shareability (through understandable representations and justifications), and ultimately makes models more likely to be translated safely into practice.

Throughout this paper, we draw on examples from recent literature to show how these principles can be operationalized. We focus especially on survival analysis and prognostic modeling, where the clinical stakes are high and model transparency is essential. Our goal is to encourage a shift in the machine learning community toward methods and practices that are not only accurate, but also interpretable, shareable, reproducible, and accountable by construction.

\section{Interpretability as a Core Foundation}
Interpretability is the cornerstone of machine learning in medicine because it enables clinicians and researchers to understand why a model makes a particular prediction. In high-stakes tasks like survival prediction, an interpretable model can highlight which variables (such as specific biomarkers, vital signs, or demographic factors) are driving a given patient’s risk estimation. This transparency is essential for building trust and for integrating model outputs into the clinical decision-making process. Moreover, interpretable models allow us to verify that the model’s reasoning aligns with medical knowledge or to discover new insights (for example, identifying a gene pathway associated with poor prognosis). In contrast, a black box model might achieve high accuracy but could be leveraging spurious correlations or biases that remain undetected until they lead to errors in practice~\cite{ref5}.

Two broad strategies exist for achieving interpretability: Post hoc explanation methods and intrinsically interpretable models. Post hoc approaches (such as LIME and SHAP) treat an already-trained model as a given and attempt to explain its predictions locally. While useful, these explanations are approximations and can be inconsistent or misleading~\cite{ref5}. In high-stakes settings, a more principled approach is to design models that are interpretable by their very structure. Classic examples in medicine include simple scoring systems or rule-based models that physicians can manually compute, but such simplistic models may lack accuracy on complex datasets. The challenge is to develop models that balance interpretability with predictive performance on modern high-dimensional data (omics, imaging, and so forth). Below, we discuss several promising directions, including traditional kernel methods, sparsity constraints, prototype-based learning, and deep kernel models, that form an interpretable toolkit for high-stakes machine learning.

\subsection{Interpretable Modeling with Traditional Kernel Methods}
Kernel methods provide a flexible framework for learning from complex data while preserving a degree of transparency. At their core, kernel machines (such as support vector machines and kernelized regressors) operate by comparing data points via a kernel function, effectively embedding data into a high-dimensional feature space. One interpretability advantage of kernel methods is that we can often design the kernel to reflect prior knowledge (e.g., via multiple kernel learning, or MKL~\cite{bach2004mkl, benhur2005ppi, damoulas2008probmkl, gonen2011mkl}). For instance, in biomedical applications we might use separate kernels for different data modalities or feature subsets (genes from different pathways, clinical vs.\ imaging data) and then combine them. MKL extends this idea by learning an optimal combination of predefined kernels, each corresponding to a specific view of the data. The learned weights on these kernels indicate which sources of information are most predictive, providing a form of interpretability at the level of data types or feature groups.

Mathematically, a typical multiple kernel learning (MKL) objective for regression or classification can be expressed as:
\[
\min_{\{\beta_m\},\,f_m \in \mathcal{H}_m} \sum_{i=1}^n L\left(y_i, \sum_{m=1}^M \beta_m f_m(x_i)\right) + \Omega(\{\beta_m\}, \{f_m\})\,,
\tag{1}
\]
where:
\begin{itemize}
  \item $L(\cdot,\cdot)$ is a loss function (e.g., squared loss for regression, hinge loss for classification),
  \item $f_m \in \mathcal{H}_m$ are functions from the RKHS induced by the $m$-th kernel $K_m$,
  \item $\beta_m \geq 0$ are non-negative combination weights for each kernel component,
  \item $\Omega(\{\beta_m\}, \{f_m\})$ is a regularization term, often of the form $\lambda \sum_{m=1}^M \beta_m \|f_m\|_{\mathcal{H}_m}^2$, promoting sparsity or smoothness.
\end{itemize}

Alternatively, in dualized kernel form (e.g., for kernel ridge regression or SVMs), the combined kernel is:
\[
K(x_i, x_j) = \sum_{m=1}^M \beta_m K_m(x_i, x_j),
\]
and optimization proceeds over $\{\beta_m\}$ subject to $\beta_m \geq 0$, often with $\sum_{m=1}^M \beta_m = 1$ to ensure identifiability. Interpreting which kernel components receive large weights $\beta_m$ can provide insight into which data modalities or feature groups are most predictive.

In prior work, an MKL-based algorithm used pathway-specific kernels to predict tumor volume from gene expression data~\cite{ref6}. Instead of hand-picking a few pathways, it simultaneously learned which gene sets (pathways) are relevant and built a predictive model, effectively performing embedded feature selection at the pathway level. The result was at least as accurate as other approaches such as random forests, while using far fewer features and providing insight into which biological pathways drive tumor growth. This interpretability at the pathway level is critical: Clinicians and biologists can examine the highlighted pathways and potentially discover mechanisms of disease progression, rather than simply receiving an opaque risk score.

A known drawback of kernel methods is their computational cost on large data sets, since naïve implementations require storing kernel matrices of size $N \times N$. However, recent advances address this issue, making kernel approaches viable for large-scale biomedical data. For instance, MAKL (Multiple Approximate Kernel Learning) integrated a kernel approximation technique, random Fourier features, with group Lasso-based feature selection in an MKL framework~\cite{ref6}. By approximating kernels, MAKL avoids explicit quadratic complexity, and the group lasso induces sparsity in the selection of kernel components. In experiments on cancer genomics data with thousands of patients, MAKL achieved state-of-the-art accuracy while selecting only a small fraction of features, yielding a sparse and interpretable model representation~\cite{ref7}. The selected kernel components (each tied to a subset of genes or features) indicate the most informative biological signals for the prediction task, aligning model interpretation with domain knowledge. These results are promising such that kernel methods, augmented with modern computational tricks, can serve as interpretable alternatives to deep neural networks, especially in scenarios where integrating heterogeneous data sources or prior knowledge is important.

Beyond MKL, other kernel-based approaches also lend themselves to interpretation. For example, Gaussian Process (GP) models use kernels to define a prior over functions and yield uncertainty estimates along with predictions. In a medical context, a GP with a Matérn or rational quadratic kernel might reveal the smoothness or characteristic length scales by which data points influence each other, offering clues about the complexity of the relationship between covariates and outcomes. Because GPs are Bayesian, they provide posterior intervals for predictions, which aligns with the need for expressing uncertainty in clinical settings. While a vanilla GP is not necessarily sparse, one can employ sparse Gaussian process techniques or couple GPs with automatic relevance determination (ARD) to infer which features have the most impact via length-scale parameters.

\subsection{Sparsity Constraints for Transparent Models}
Imposing sparsity is a classical and powerful technique to enhance interpretability. Sparse models intentionally use only a small subset of the available features (or basis functions) for prediction, making it easier for a human to track the model’s reasoning. One canonical example is the Lasso (L$_1$-regularized) linear model~\cite{ref8}, which performs feature selection by driving many coefficients to zero. In medical applications, a Lasso-based Cox proportional hazards model can select a handful of prognostic factors from a high-dimensional set of covariates (such as gene expression levels) while providing a linear risk score that experts can interpret. To recall, the partial likelihood for a Cox model with parameter vector $\beta$ is often written as:
\[
L(\beta) = \prod_{i=1}^n \frac{\exp(\beta^\top x_i)}{\sum_{j \in R_i} \exp(\beta^\top x_j)}^{\delta_i}\!,
\tag{2}
\] 
where $R_i$ is the risk set for subject $i$ and $\delta_i$ indicates whether the subject’s event is observed. Adding an L$_1$ penalty $\|\beta\|_1$ leads to sparsity in $\beta$, which yields interpretability in terms of which covariates have nonzero coefficients.

Sparsity can also be imposed at the group level, such as when features have a natural grouping into pathways or measurement clusters. This is useful in genomics, where multiple measurements may belong to the same biological process. In methods such as MAKL, group sparsity is achieved by treating each approximate kernel matrix (associated with a feature group) as a single unit to be selected or not. The result is that only a few kernel components are given nonzero weight, greatly simplifying the model’s explanation: It might say that “the model bases its predictions primarily on pathway A and pathway B,” which is easily interpretable by domain experts.

Rule-based or tree-based models can also benefit from sparsity. A shallow decision tree can yield a small set of rules, each involving only a handful of features. Ensemble methods like random forests are less interpretable in raw form, but one can extract a small set of representative rules or measure overall feature importance across the ensemble. However, caution is needed: Simple importance measures, like the Gini importance in random forests, can be misleading in the presence of correlated variables. Sparse tree methods (such as a tree with depth constraints) or rule extraction with regularization can produce more faithful explanations.

Sparsity can be introduced in deep networks through attention mechanisms or specialized regularizers that encourage the network to focus on a smaller subset of inputs or hidden units. Although deep learning is often associated with thousands or millions of parameters, local sparsity patterns may make interpretability more tractable. For instance, a survival prediction network might be designed so that its attention on input features is sparse, highlighting only the most relevant clinical factors for each patient. Research on mask learning attempts to discover binary masks within neural networks that directly reveal which features or neurons are active in making predictions. While the full network may still be complex, these masks can provide a partial window into the model’s logic.

\subsection{Prototype-Based Learning and Case-Based Reasoning}
Another intuitive form of interpretability is prototype-based reasoning, where a model explains its predictions by reference to representative examples in the data. This approach resonates with clinicians, who often recall prototypical cases when making judgments. 

Recent research in interpretable deep learning extends this idea using prototype layers that learn a small set of prototypical representations. At inference time, a new instance is compared to these prototypes, and the prediction is expressed as a weighted combination of the similarities. Chen \emph{et al.}~\cite{ref9} illustrated this in the context of image analysis, where the network highlights which prototypical image parts resemble the query. In medicine, one could imagine prototypes corresponding to typical patterns of disease progression or characteristic genomic profiles. A survival model might say that “patient X is similar to prototype 1 (an observed patient with short survival) and prototype 2 (a patient with moderate survival), hence we predict an intermediate risk.” By grounding predictions in concrete examples, clinicians can see tangible evidence supporting each risk assessment. Indeed, prototype-based methods have been explored in clinical settings (e.g., medical imaging) to provide case-based explanations for model decisions~\cite{ref10}.

Prototype methods often rely on a dictionary learning mechanism or a clustering step. They can also incorporate sparsity by learning a small number of prototypes. In a clinical setting, it is beneficial if prototypes correspond to real patients, so that an actual medical record can be displayed. If synthetic prototypes are used, they should still be interpretable (for instance, a synthetic set of lab values that are within plausible ranges). Such case-based reasoning naturally complements feature-based approaches like Lasso, because it helps answer not just “which features matter?” but also “which past patients (or typical patterns) does this case resemble?”

\subsection{Interpretable Deep Learning and Deep Kernel Models}
Deep learning approaches for medical data often face criticism for their lack of transparency: their hidden layers and nonlinearities can obscure how predictions are formed. Nevertheless, there is a growing push toward interpretable deep learning. One strategy is to impose constraints on the network architecture, such as in neural additive models~\cite{ref11}, where each input variable contributes via a learned univariate or low-dimensional function that can be visualized. Another strategy is concept bottleneck models, where intermediate features correspond to human-recognizable concepts~\cite{ref12}. 

Attention mechanisms can also reveal which portions of an input sequence the model focuses on, especially if the attention is designed to be sparse. Coupling deep networks with kernels is another promising direction. Deep kernel learning~\cite{ref13} uses a neural net to map inputs to an embedding, and then applies a Gaussian kernel (or related kernel) in that embedding space. Mathematically:
\[K_{\text{DL}}(x_i, x_j) = K\!\big(f_\theta(x_i),\,f_\theta(x_j)\big)\,,\tag{3}\] 
where $f_\theta$ is a learned neural feature extractor. This merges the advantages of neural representation learning with the interpretability tools of kernel machines, such as support vectors or Gaussian processes with uncertainty estimates. 
Finally, some deep networks incorporate prototype or case-based layers, aligning the deep learned features with example-driven explanations. Alternatively, one can perform a post hoc analysis of the embedding learned by a deep net to find representative prototypes in that latent space. Although interpretability in deep learning remains a challenge, these approaches illustrate that purely opaque black boxes are not inevitable. By designing architectures with interpretability objectives, or by combining deep learning with kernel or prototype ideas, it becomes feasible to leverage deep models for high-stakes medical tasks without sacrificing trust and transparency.

\subsection{Comparison of Interpretability Methods}
We summarize several interpretability strategies in Table~1, comparing their explanation styles and key advantages.

\begin{table}[ht]
\centering
\caption{Interpretability methods for clinical machine learning: explanation style and practical advantages}
\resizebox{\textwidth}{!}{%
\begin{tabular}{llll}
\toprule
\textbf{Method} & \textbf{Explanation Style} & \textbf{Key Advantage}  \\
\midrule
Kernel Methods (e.g., MKL)       & Feature-level (via kernel weights) & Integrates prior knowledge; modular  \\
Sparsity-based Models           & Sparse feature coefficients        & Simple model structure               \\
Prototype-based                 & Case-level examples               & Clinically intuitive                   \\
Neural Additive Models          & Additive feature contributions    & Smooth partial dependence              \\
Post hoc Explainers             & Local attributions (e.g., SHAP)   & Broadly applicable                    \\
\bottomrule
\end{tabular}
}
\end{table}

The table highlights how kernel methods can reflect domain knowledge through carefully chosen kernels, sparsity-based models yield concise feature sets, prototype models link outputs to similar past cases, and so forth. In practice, one may combine these techniques: for instance, a multiple kernel learning model can also produce prototype-based explanations by identifying support vectors or representative points in its transformed feature space.

\section{Accountability in Model Development and Deployment}
Accountability refers to the obligation of models and their developers to ensure reliability, fairness, and compliance with ethical and regulatory standards. In medical machine learning, an accountable model is one whose performance has been thoroughly evaluated, whose limitations and uncertainties are documented, and whose use avoids reinforcing bias or causing unjust harm. Accountability is facilitated by interpretability, but also requires robust evaluation protocols that go beyond model design.

\textbf{Rigorous Validation and Reproducibility.} An accountable medical ML model must undergo rigorous validation, ideally on external data sets that differ from the training set. It is well documented that many models perform well on internal data but fail to generalize to new populations. For example, a sepsis detection model trained on one institution’s data might not replicate at another hospital with different demographics~\cite{ref16}. Testing models on public benchmarks can reveal overfitting and enhance reproducibility. In survival analysis, verification of model performance on external cohorts is essential. When data cannot be shared, techniques described in Section~4 (such as synthetic data generation or federated testing) can facilitate external validation without compromising privacy. Reproducibility also requires that code, hyperparameters, and relevant metadata are properly documented or shared, allowing other investigators to confirm the results.
\subsection*{Appropriate Metrics and Uncertainty Quantification}
Choosing metrics that reflect the clinical objectives is crucial. In survival analysis, the concordance index (C-index) is often used to measure discriminative ability, indicating whether higher risk predictions correlate with earlier events. This was introduced and discussed earlier by~\cite{harrell1982, harrell1984, gonenheller2005} , and can be expressed as:

\begin{equation}
\widehat{C} = 
\frac{
\sum_{i,j} \mathbb{I}(\hat{r}_i > \hat{r}_j)\,\mathbb{I}(T_i < T_j)\,\delta_i
}{
\sum_{i,j} \mathbb{I}(T_i < T_j)\,\delta_i
},
\tag{4}
\end{equation}

where $T_i$ is the survival time for patient $i$, $\delta_i$ is the event indicator, and $\hat{r}_i$ is the model’s risk score. A model with a higher C-index more effectively ranks patients by risk.

Calibration is also important, because a model that is discriminative but poorly calibrated might mislead clinicians with overly optimistic or pessimistic probabilities of survival. Furthermore, accountability demands uncertainty quantification, because medical decisions often hinge on the confidence in a risk score. Methods like Bayesian neural networks, Gaussian processes, or bootstrapping can provide intervals or distributions for predictions, enabling clinicians to appreciate when the model might be uncertain.

A single point estimate rarely suffices in high-stakes settings. Standard errors or confidence intervals, often derived via U-statistics or bootstrapping, help convey how robust the estimated C-index is~\cite{harrell1982,gonenheller2005}. Complementary measures, like the integrated Brier score, capture aspects of calibration and predictive accuracy over time~\cite{schemper1996explained}. Indeed, even if a model ranks patients well, it must also provide reliable risk estimates or survival probabilities for effective clinical decision making. Hence, broader accountability demands reporting metrics that collectively demonstrate discrimination, calibration, and uncertainty, regardless of whether the underlying task is survival prediction, classification, or a related problem.

\textbf{Fairness and Bias Mitigation.} Accountable ML systems must be examined for potential biases. If the training data underrepresent certain demographics, the model may exhibit disparate performance across groups. Bias can also arise if the model indirectly uses proxies for protected attributes. Researchers should audit performance by subgroups (for example, by race, age, or gender) to identify gaps. If significant disparities appear, mitigation can involve rebalancing the training set or modifying the objective to enforce fairness constraints on error rates or calibration. Interpretability can further help identify features that cause spurious or unfair distinctions, revealing whether the model is learning an unintentional proxy for sensitive attributes. Such biases are not merely hypothetical: an analysis of a widely used health risk algorithm revealed substantial racial bias in its predictions~\cite{ref17}.

\textbf{Model Monitoring and Updates.} Accountability extends into the deployment phase, where model performance can drift over time due to changes in clinical practice or shifts in population. Continuous monitoring and periodic model updates address this challenge. For example, a survival model may need retraining as treatment protocols evolve, or as new biomarkers become relevant. Institutions that deploy AI in production should maintain logs of predictions and outcomes, enabling retrospective analyses of mistakes or biases that might emerge. Transparency about known limitations is also part of accountability: When a survival model is less reliable for certain rare conditions or specific subpopulations, that should be disclosed to the end users (clinicians, patients, or regulatory bodies). Model cards~\cite{ref18} or similar documentation serve to communicate these details clearly.

In summary, accountability in medical ML requires diligence throughout the entire life cycle: From design (ensuring interpretability), to validation (with domain-appropriate metrics and external data), to real-world deployment (with fairness audits, uncertainty quantification, and continuous updates). These practices strengthen confidence in AI-driven tools for critical decisions such as cancer screening or therapy selection.

\section{Shareability, Reproducibility, and Collaboration via Generative and Federated Approaches}
Medical data are often sensitive, siloed, and heterogeneous, which poses significant hurdles to developing robust, generalizable machine learning models. Shareability refers to the ability of researchers and clinicians to share either data or knowledge without violating patient privacy or institutional restrictions. This dimension is vital for reproducibility, since external validation of results is difficult if the original data cannot be seen by others. In this section, we discuss how generative AI and federated learning can facilitate the secure exchange of information, reduce redundancy, and enhance collaborative efforts across institutions.

\subsection{Generative AI for Data Augmentation and Synthetic Data Sharing}
Generative modeling has shown impressive capabilities in synthesizing realistic yet artificial samples that resemble real patients. Frameworks such as generative adversarial networks (GANs), variational autoencoders (VAEs), and diffusion models can capture high-dimensional distributions of images or tabular data. For instance, Choi \emph{et al.}~\cite{ref19} proposed medGAN to generate discrete patient records, showing that synthetic data can preserve many statistical properties of the original data while obscuring individual identities. In survival analysis, diffusion- or GAN-based methods could generate patient trajectories (covariates over time plus an event time), aiding research on scarce conditions or on balancing data sets with rare patient types.

When data sets cannot be shared directly, synthetic data can serve as a proxy for validating algorithms or exploring hypotheses. Researchers at other sites can download the synthetic data, train or test models, and see if their conclusions align with those obtained by the original team. This addresses reproducibility in part, although it is important to confirm that the synthetic data truly capture the relevant real-data characteristics. Additionally, generative models can be used for domain adaptation or filling in missing modalities, by learning cross-modality mappings (for instance, inferring lab values given imaging features).

Nevertheless, generative approaches must be used cautiously. If the generative model overfits its training set, it may leak private details about real patients. Differential privacy methods can mitigate this risk by injecting noise during training, ensuring that no single individual’s data can be precisely recovered~\cite{ref20, ref21}. Achieving high fidelity while preserving privacy is an active research area. Provided these concerns are addressed, synthetic data stands to greatly advance shareability and reproducibility in medical ML, since investigators in different locations can exchange knowledge without sharing protected health information.

\subsection{Foundation Models and Cross-Institutional Integration}
Foundation models are large-scale models pre-trained on diverse data sets that can be adapted to a wide range of tasks, often by fine-tuning. In the biomedical domain, examples include large language models trained on scientific text and electronic health records, or multimodal models that incorporate medical imaging as well. These models, once trained, can be shared with collaborators as a compact representation of knowledge gathered from multiple institutions~\cite{ref22}. Instead of distributing raw data, one can distribute the model weights, which are presumably less sensitive (though not entirely free of privacy concerns). 

Such foundation models can facilitate cross-institutional integration because they often learn features or embeddings that generalize across populations. A hospital with limited data might fine-tune a foundation model on its local data, obtaining good performance without needing a huge local training set. Meanwhile, knowledge from other institutions is indirectly embedded in the model’s parameters. This approach has been explored for tasks such as radiology (e.g., detecting tumors on MRI scans), pathology (classifying tissue images), and analyzing clinical notes~\cite{ref22}. Model performance is monitored for biases or domain shifts, and if issues arise, further fine-tuning or domain adaptation can be performed.

\subsection{Federated Learning and Privacy-Preserving Collaboration}
Federated learning (FL) is another powerful paradigm for enabling collaboration among multiple institutions without centralizing raw data~\cite{ref23}. In FL, each site trains the model locally on its proprietary data, and only the aggregated weight updates or parameter gradients are shared. This allows a global model to be learned, reflecting information from all sites, while never exposing any individual patient’s data. Federated learning has been successfully applied in COVID-19 outcome prediction~\cite{ref24}, brain tumor segmentation from MRI scans~\cite{ref25}, and other multi-center scenarios (e.g., across hospitals and research labs) where data pooling is infeasible.

FL can incorporate differential privacy~\cite{ref20} or secure multi-party computation to further reduce the risk of information leakage. This synergy helps address legal and ethical constraints that often prevent data sharing. FL also naturally promotes multi-site validation, because each site can measure performance on its own data set. Furthermore, interpretability methods can be federated by design: for instance, a federated sparse linear model can reveal which features are generally important across sites, or site-specific differences might emerge in the model coefficients.

Despite these advantages, FL is not without challenges. Heterogeneity among institutional data (variations in feature availability, patient populations, and measurement protocols) can complicate model aggregation. Communication overhead can increase with large models, although strategies such as secure parameter compression help. Also, ensuring accountability may require that each site reviews local updates and audits potential biases. Nonetheless, federated learning stands out as a promising approach to building robust, shareable models that transcend institutional silos.

\section{Conclusion}
In this position paper, we have argued that interpretability, shareability, reproducibility and accountability should be first-class objectives, on par with predictive accuracy, when designing machine learning systems for high-stakes medical applications. We surveyed methodological approaches (kernel methods, sparsity, prototype-based models, deep kernels) and supporting frameworks (federated learning, generative data sharing) that can enable more transparent, trustworthy, and reproducible models.

It is our view that, when deploying models to guide real clinical decisions, an interpretable approach (or a hybrid that ensures interpretability) should be the default. While black box models may still be beneficial in certain research contexts, practical and regulatory concerns favor architectures that can provide human-understandable explanations. In high-stakes settings, interpretable models are preferable to black box models plus post hoc explanations~\cite{rudin2019stop}.

Accountability further requires that we adopt proper metrics, carefully analyze uncertainty, and openly disclose performance across different patient strata, while continuing to monitor models in routine use. In survival analysis, the role of adequate censoring-aware metrics and external validation cannot be overstated.

Finally, generative modeling and federated learning open the door for cross-institutional collaboration without direct data sharing, addressing a central barrier to reproducibility in medical ML. Foundation models, if carefully governed, may encapsulate collective knowledge gleaned from disparate sites, accelerating progress without violating patient privacy.

In essence, we believe the machine learning community should champion interpretability, shareability, reproducibility and accountability as fundamental aspects of medical AI. By doing so, we can create models that earn the trust of clinicians and patients, and that genuinely improve health outcomes rather than merely optimizing narrow performance metrics on internal data sets. We hope this paper stimulates continued research and discussion on how to integrate these aspects into the core of biomedical data science.

\end{document}